\documentstyle[12pt]{article}

\newlength{\smpagewidth}
\newlength{\smpageheight}

\newcommand{\setleftmargin}[1]{
	\addtolength{\textwidth}{\oddsidemargin}
	\addtolength{\textwidth}{1in}
	\addtolength{\textwidth}{-#1}
	\setlength{\oddsidemargin}{-1in}
	\addtolength{\oddsidemargin}{#1}
	\setlength{\evensidemargin}{\oddsidemargin}
}
\newcommand{\setrightmargin}[1]{
	\setlength{\textwidth}{\smpagewidth}
	\addtolength{\textwidth}{-\oddsidemargin}
	\addtolength{\textwidth}{-1in}
	\addtolength{\textwidth}{-#1}
}
\newcommand{\settopmargin}[1]{
	\addtolength{\textheight}{\topmargin}
	\addtolength{\textheight}{1in}
	\addtolength{\textheight}{\headheight}
	\addtolength{\textheight}{\headsep}
	\addtolength{\textheight}{-#1}
	\setlength{\topmargin}{-1in}
	\addtolength{\topmargin}{-\headheight}
	\addtolength{\topmargin}{-\headsep}
	\addtolength{\topmargin}{#1}
}
\newcommand{\setbottommargin}[1]{
	\setlength{\textheight}{\smpageheight}
	\addtolength{\textheight}{-\topmargin}
	\addtolength{\textheight}{-1in}
	\addtolength{\textheight}{-\footskip}
	\addtolength{\textheight}{-#1}
}

\setleftmargin{3cm}
\setrightmargin{3cm}
\settopmargin{2.2cm}
\setbottommargin{2.5cm}

\newcommand{\singlespace}{}
\newcommand{\shortcite}[1]{\cite{#1}}

\author{Michael White\thanks{Most of this research was undertaken
at the University of Pennsylvania, where the author was partially
supported by an NSF Graduate Fellowship.  For helpful comments
and discussion, the author thanks Mark Steedman, Richard Oehrle,
Alessandro Zucchi, Barbara Di Eugenio, Matthew Stone, Christine Doran,
Owen Rambow and two anonymous reviewers.} \\
CoGenTex, Inc. \\
The Village Green \\
840 Hanshaw Road \\
Ithaca, NY 14850, USA \\
{\tt mike@cogentex.com }}

\title{A Computational Approach to \\ Aspectual Composition \\
\mbox{} \\
{\it Une perspective computationelle \\
sur la composition aspectuelle} }

\date{}

\begin{document}

\maketitle

\pagestyle{empty}
\thispagestyle{empty}

\begin{abstract}
\noindent
In this paper, I argue, contrary to the prevailing opinion in the
linguistics and philosophy literature, that a sortal approach to
aspectual composition can indeed be
explanatory.  In support of this view, I develop a synthesis of
competing proposals by Hinrichs, Krifka and Jackendoff which takes
Jackendoff's cross-cutting sortal distinctions as its point of
departure.  To show that the account is well-suited for computational
purposes, I also sketch an implemented calculus of eventualities which
yields many of the desired inferences.  Further details on both the
model-theoretic semantics and the implementation can be found in
\cite{White94Thesis}.
\mbox{} \\
\mbox{} \\
\noindent
{\it
Dans cet article, je propose, contrairement aux opinions
pr\'{e}pond\'{e}rantes dans la litt\'{e}rature linguistique et
philosophique, qu'une theorie ``sortale'' de la composition
aspectuelle peut \^{e}tre explicative.  Comme justification de cette
th\`{e}se, je d\'{e}veloppe une synth\`{e}se des th\'{e}ories en
comp\'{e}tition de Hinrichs, de Krifka et de Jackendoff, qui prend
comme point de d\'{e}part les distinctions sortales
multi-dimensionelles de Jackendoff.  Afin de montrer que la solution
pr\'{e}sent\'{e}e peut \^{e}tre exploit\'{e}e computationellement,
j'esquisse un calcul d'\'{e}ventualite's impl\'{e}ment\'{e} qui permet
de faire beaucoup des inf\'{e}rences desir\'{e}es.  Des d\'{e}tails
suppl\'{e}mentaires sur la s\'{e}mantique \`{a} base de mod\`{e}les et
sur l'impl\'{e}mentation peuvent
\^{e}tre trouv\'{e}s dans \cite{White94Thesis}.
}
\end{abstract}

\vspace{1cm}

\section{Introduction}

In recent years, it has become common in the linguistics and
philosophy literature to assume that {\bf events} and {\bf processes}
are ontologically distinct entities, on a par with {\bf objects} and
{\bf substances}.  At the same time, the idea that episodic knowledge
should be represented as a collection of interrelated {\bf
eventualities} has gained increasing acceptance in the computational
linguistics and artificial intelligence literature.

Contrary to what one might expect, a search through the prior
literature in linguistics and philosophy reveals no account in which
these sortal distinctions play a direct role in adequately explaining
the problem of {\bf aspectual composition} \cite{Dowty72,Verkuyl72}.
In fact, amongst those that have explicitly considered this question,
the consensus appears to be that no such explanation is likely to be
found \cite{Krifka92,Moltmann91,Verkuyl93}.  From a computational
perspective this is rather unfortunate, since such distinctions have
otherwise proved quite useful
\cite{Dale92,Eberle92,HwangSchubert92,LascaridesEtAl92}.

In this paper, I set out to show that a sortal approach to aspectual
composition, developed in the spirit of the eventuality-based work on
episodic representation, can indeed be explanatory.  In so doing, I
develop a synthesis of competing proposals by
Hinrichs~\shortcite{Hinrichs85}, Krifka~\shortcite{Krifka92} and
Jackendoff~\shortcite{Jackendoff91} which takes Jackendoff's
cross-cutting sortal distinctions as its point of departure.  To show
that the account is well-suited for computational purposes, I also
sketch an implemented calculus of eventualities which yields many of
the desired inferences.  Further details on both
the model-theoretic semantics and the implementation can be found in
\cite{White94Thesis}.

\section{Motivation}

{}From a knowledge-representation perspective, eventuality-based
representations have proven to be convenient for their conciseness,
support for underspecificity, and easy integration with natural
language interfaces.  For example, consider the following
attribute-value representation of a {\em filling\/} event
$e_0$ (the representations in this section are meant to be
reminiscent of the ones in Dale~\shortcite{Dale92}, which are in turn based
largely upon Bach~\shortcite{Bach86}):

{\singlespace
\[
\begin{array}{c}
\left[
\begin{array}{l}
{\sf index}: \; e_0 \\
{\sf sort}: \; {\sf event} \\
{\sf pred}: \; {\sf fill} \\
{\sf agent}: \; {\sf jack} \\
{\sf patient}: \;
  \left[
  \begin{array}{l}
  {\sf index}: \; x_0 \\
  {\sf sort}: \; {\sf object} \\
  {\sf pred}: \; {\sf bucket} \\
  {\sf card}: 5 \\
  \end{array}
  \right] \\
{\sf duration}:
  \left[
  \begin{array}{l}
  {\sf number}: 20 \\
  {\sf unit}: \; {\sf minutes} \\
  \end{array}
  \right] \\
\end{array}
\right] \\
\mbox{} \\
\mbox{\em Jack filled five buckets in twenty minutes} \\
\end{array}
\]
}

\noindent It should be evident that this structure can be
straightforwardly translated into the English sentence {\em Jack
filled five buckets in twenty minutes}.\footnote{For simplicity I
shall ignore the encoding of tense in this paper.} To see that this
representation is also concise, note that one can also derive numerous
other sentences from this structure, given appropriate rules of
inference: for example, {\em Jack filled a bucket}, {\em Jack filled
something}, etc.  On the other hand, this representation also supports
underspecificity, since from this structure one cannot determine which
five buckets were filled (e.g., bucket A, \ldots, bucket E).

As a second example, let us now consider the following representation
of a {\em pouring\/} process $e_1$:

{\singlespace
\[
\begin{array}{c}
\left[
\begin{array}{l}
{\sf index}: \; e_1 \\
{\sf sort}: \; {\sf process} \\
{\sf pred}: \; {\sf pour} \\
{\sf agent}: \; {\sf jack} \\
{\sf patient}: \;
  \left[
  \begin{array}{l}
  {\sf index}: \; x_1 \\
  {\sf sort}: \; {\sf substance} \\
  {\sf pred}: \; {\sf water} \\
  \end{array}
  \right] \\
{\sf goal}: \;
  \left[
  \begin{array}{l}
  {\sf index}: \; x_a \\
  {\sf sort}: \; {\sf object} \\
  {\sf pred}: \; {\sf bucket} \\
  {\sf name}: \; {\sf A} \\
  \end{array}
  \right] \\
{\sf duration}:
  \left[
  \begin{array}{l}
  {\sf number}: 30 \\
  {\sf unit}: \; {\sf seconds} \\
  \end{array}
  \right] \\
\end{array}
\right] \\
\mbox{} \\
\mbox{\em Jack poured water into bucket A for thirty seconds} \\
\end{array}
\]
}

\noindent It should be evident once again that this
structure can be straightforwardly translated into the English
sentence {\em Jack poured water into bucket A for thirty seconds}.
Moreover, this representation is similarly concise: given appropriate
rules of inference, one can also derive {\em Jack poured water into
bucket A for twenty-five seconds}, {\em Jack poured water into bucket
A for twenty seconds}, and so forth.\footnote{While these sentences
are clearly not as informative, this is a matter of implicature; to
see this, contrast these sentences with their counterparts in the
previous case (i.e., {\em Jack filled five buckets in
fifteen/ten/\ldots minutes}), which need not be true.} Finally, this
representation likewise supports underspecificity, since in the
absence of any information about the rate of transfer one cannot
determine how much water was poured into bucket A.

As the careful reader may have noticed, the choice of temporal
adverbial in the preceding examples is conditioned by the sort of
eventuality in question, which again depends on the verb.  This is not
the whole story, however: given a particular amount of water
in the second example, the appropriate adverbial changes (cf. {\em
Jack poured five gallons of water into bucket A *for/in
thirty seconds}); contrariwise, switching to a bare plural in the
first example, we may note a switch in the opposite direction (cf.
{\em Jack filled buckets for/*in twenty minutes}).  Adequately
explaining dependencies such as these is the problem of {\bf aspectual
composition}, to which we now turn.

\section{A Sortal Approach to Aspectual Composition}

To make the ensuing discussion more concrete, let us examine the
following possible representation for the sentence {\em Jack poured
five gallons of water into bucket A in thirty seconds\/} (as in the
previous section, this representation is meant to be reminiscent of
those found in \cite{Dale92}):

{\singlespace
\[
\begin{array}{c}
\left[
\begin{array}{l}
{\sf index}: \; {e_1}^\prime \\
{\sf sort}: \; {\sf event} \\
{\sf pred}: \; {\sf pour} \\
{\sf agent}: \; {\sf jack} \\
{\sf patient}: \;
  \left[
  \begin{array}{l}
  {\sf index}: \; {x_1}^\prime \\
  {\sf sort}: \; {\sf object} \\
  {\sf pred}: \; {\sf water} \\
  {\sf quantity}: \;
    \left[
    \begin{array}{l}
    {\sf number}: \; 5 \\
    {\sf unit}: \; {\sf gallons} \\
    \end{array}
    \right] \\
  \end{array}
  \right] \\
{\sf goal}: \;
  \left[
  \begin{array}{l}
  {\sf index}: \; x_a \\
  {\sf sort}: \; {\sf object} \\
  {\sf pred}: \; {\sf bucket} \\
  {\sf name}: \; {\sf A} \\
  \end{array}
  \right] \\
{\sf duration}:
  \left[
  \begin{array}{l}
  {\sf number}: 30 \\
  {\sf unit}: \; {\sf seconds} \\
  \end{array}
  \right] \\
\end{array}
\right] \\
\mbox{} \\
\mbox{\em Jack poured five gallons of water into bucket A
in thirty seconds} \\
\end{array}
\]
}

\noindent Comparing this representation to the previous one (for
{\em Jack poured water into bucket A for thirty seconds}), two
questions naturally arise:

\begin{itemize}
\item Why has $e_1$ changed to ${e_1}^\prime$? (And
$x_1$ to ${x_1}^\prime$?)

\item Assuming this has something to do with their
differing sortal values,
why should the the sort of ${e_1}^\prime$ depend upon
that of ${x_1}^\prime$ in the first place?
\end{itemize}

Remarking on a question similar to the first one (and translating
their remarks into the present context), Oberlander and
Dale~\shortcite{OberlanderDale91} observe that the primed and unprimed
entities cannot be the same, since their respective sorts are assumed
to be disjoint.  They then suggest that this is in some sense to be
expected, asserting that the two sentences these representations give
rise to convey different {\em perspectives\/} on the same situation;
at the same time though, they also acknowledge that at some level we
would like to tie these two (supposed) perspectives together.

Interestingly, a similar problem arises with
Jackendoff's~\shortcite{Jackendoff91} conception of the binary feature
$\pm{\sf b(ounded)}$, which he introduces to distinguish both events
($+{\sf b}$) from processes ($-{\sf b}$) and objects ($+{\sf b}$) from
substances ($-{\sf b}$).  According to Jackendoff, ``a speaker uses a
$-{\sf b}$ constituent to refer to an entity whose boundaries are not
in view or not of concern; one can think of the boundaries as outside
the current field of view.''  Although this idea has some appeal when one
focuses on the discourse-backgrounding function that atelic sentences
can have, it does not seem particularly apt here, where our view of
what has taken place remains constant.

While our two example sentences clearly convey different information,
it is not at all obvious how to make sense of this difference in terms
of perspectives.  For this reason, I will pursue an alternative
approach below which obviates the need to do so.  As we shall see,
this will require us to develop an alternative conception of the
event/process (and object/substance) distinction than Jackendoff
appears to have in mind.

Turning now to the second question, we should first note that {\em
Jack poured water into bucket A\/} and {\em Jack filled bucket A with
water\/} exhibit different aspectual behavior, despite the presence of
the same mass noun.  According to traditional wisdom, the relevant
difference between {\em fill\/} and {\em pour\/} is that only the
former encodes what Krifka~\shortcite{Krifka92} calls a {\bf set
terminal point}: if an event $e$ is a filling event, then (presumably)
no proper part of $e$ is also a filling event, and thus {\em fill\/}
is taken to encode a set terminal point; since this argument does not
(presumably) go through for pouring events, {\em pour\/} is not taken
to encode a set terminal point.  Of course, while {\em pour\/} itself
does not supply a set terminal point, one may be supplied indirectly
by specifying a fixed quantity of what is poured; in this case, the
terminal point coincides with the eventual exhaustion of this
quantity.

As Krifka points out, the conventional wisdom about aspectual composition
does not appear to be compatible with a sortal approach:

\begin{quote}
{\singlespace
For consider a concrete event of running and a concrete event of
running a mile; then surely both events have a terminal point
(both events might even be identical).  The difference is that
an event of running might be part of another event of running which
has a later terminal point, whereas this is not possible for an event
of running a mile.
}
\end{quote}

\noindent For this reason, Krifka~\shortcite{Krifka92} eschews
sortal distinctions (amongst eventualities) and develops an account
based upon the reference properties of event predicates instead.
Unfortunately though, this decision leads to empirical problems with
{\bf non-individuating accomplishment expressions}, such as {\em run
more than a mile} (\cite{White94Thesis}; cf. also
Verkuyl~\shortcite{Verkuyl93}): grammatically, {\em run more than a
mile\/} patterns with {\em run a mile}, yet according to Krifka's
test, {\em run more than a mile} patterns with {\em run} (note that an
event of running more than a mile might be part of another event of
running more than a mile).

What Krifka's observation suggests is that in pursuing a sortal approach,
we should look for alternatives to the sortal distinctions assumed
so far, which are based upon Bach~\shortcite{Bach86}.  One possibility
is to assume that substances and processes are more abstract entities
than objects and events:  for example, rather than letting a substance
be a particular quantity of matter, we may assume a substance is a
{\em continuum\/} of such quantities; likewise, we may take a process
to be a continuum of events with differing durations.

To relate these continua to their particulars, I will borrow
Jackendoff's {\sf composed-of} relation (though not necessarily its
original semantics).  Following Jackendoff, I will assume that this
relation forms part of the meaning of {\bf measure phrases}, which
include adjectival ones such as {\em five gallons of\/} and adverbial
ones such as {\em for thirty seconds}.\footnote{Note that this term is
not intended to include restrictive modifiers of measurement, such as
{\em in thirty seconds\/} or {\em (a) five-gallon (X)}; although this
choice of terminology appears to be consistent with that of
Moltmann~\shortcite{Moltmann91}, perhaps a happier term could be
found.}  This yields the following representations for our example
sentences:\footnote{NB:  These representations are not
the same as those found in \cite{White94Thesis}, where a more standard
logical representation is employed.  For present purposes, this
difference is assumed to be of no importance.}

{\singlespace
{\small
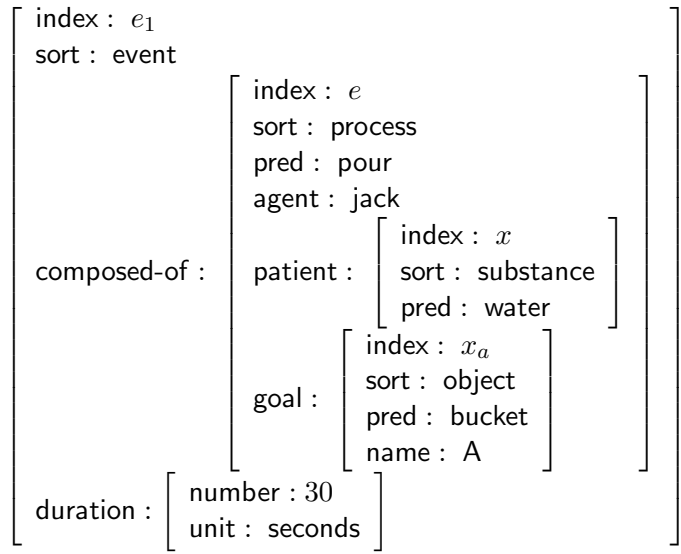
\[
\begin{array}{c}
\left[
\begin{array}{l}
{\sf index}: \; e_1 \\
{\sf sort}: \; {\sf event} \\
{\sf composed\mbox{-}of}: \;
  \left[
  \begin{array}{l}
  {\sf index}: \; e \\
  {\sf sort}: \; {\sf process} \\
  {\sf pred}: \; {\sf pour} \\
  {\sf agent}: \; {\sf jack} \\
  {\sf patient}: \;
    \left[
    \begin{array}{l}
    {\sf index}: \; x \\
    {\sf sort}: \; {\sf substance} \\
    {\sf pred}: \; {\sf water} \\
    \end{array}
    \right] \\
  {\sf goal}: \;
    \left[
    \begin{array}{l}
    {\sf index}: \; x_a \\
    {\sf sort}: \; {\sf object} \\
    {\sf pred}: \; {\sf bucket} \\
    {\sf name}: \; {\sf A} \\
    \end{array}
    \right] \\
  \end{array}
  \right] \\
{\sf duration}:
  \left[
  \begin{array}{l}
  {\sf number}: 30 \\
  {\sf unit}: \; {\sf seconds} \\
  \end{array}
  \right] \\
\end{array}
\right] \\
\mbox{} \\
\mbox{\em Jack poured water into bucket A for thirty seconds} \\
\end{array}
\]
}}

{\singlespace
{\small
\[
\begin{array}{c}
\left[
\begin{array}{l}
{\sf index}: \; e_1 \\
{\sf sort}: \; {\sf event} \\
{\sf pred}: \; {\sf pour} \\
{\sf agent}: \; {\sf jack} \\
{\sf patient}: \;
  \left[
  \begin{array}{l}
  {\sf index}: \; x_1 \\
  {\sf sort}: \; {\sf object} \\
  {\sf composed\mbox{-}of}: \;
    \left[
    \begin{array}{l}
    {\sf index}: \; x \\
    {\sf sort}: \; {\sf substance} \\
    {\sf pred}: \; {\sf water} \\
    \end{array}
    \right] \\
  {\sf quantity}: \;
    \left[
    \begin{array}{l}
    {\sf number}: \; 5 \\
    {\sf unit}: \; {\sf gallons} \\
    \end{array}
    \right] \\
  \end{array}
  \right] \\
{\sf goal}: \;
  \left[
  \begin{array}{l}
  {\sf index}: \; x_a \\
  {\sf sort}: \; {\sf object} \\
  {\sf pred}: \; {\sf bucket} \\
  {\sf name}: \; {\sf A} \\
  \end{array}
  \right] \\
{\sf duration}:
  \left[
  \begin{array}{l}
  {\sf number}: 30 \\
  {\sf unit}: \; {\sf seconds} \\
  \end{array}
  \right] \\
\end{array}
\right] \\
\mbox{} \\
\mbox{\em Jack poured five gallons of water into bucket A
in thirty seconds} \\
\end{array}
\]
}}

\noindent To paraphrase, in the first case $e_1$ is an event of
duration thirty seconds which is composed of a process $e$ in which
Jack pours the substance $x$, which is water, into bucket A.  In the
second case, $e_1$ is instead an event (again of duration thirty
seconds) in which Jack pours the object $x_1$ into bucket A, where
$x_1$ is a five-gallon quantity (composed-) of the substance $x$
(which is again water).

At this point we may answer the two questions with which we began this
section.  With respect to the first question (is it necessary to
introduce a new eventuality ${e_1}^\prime$ to represent the second of
our two sentences?), the above representations show that we can now
simply treat these two sentences as two different descriptions of the
same event $e_1$ --- much as in Krifka's treatment --- without causing
a sortal clash.

Returning now to the second question (why should we observe sortal
dependencies in the first place?), let us consider how the
conventional wisdom can be reconstructed here.  In
\cite{White94Thesis}, I suggest that the relation established by {\em
pour\/} between a material entity and an eventuality is an instance of
an {\bf incremental thematic relation}, following the terminology of
Dowty~\shortcite{Dowty91}.  What characterizes such relations is how
predication over continua is to be understood: predication should only
involve {\bf delimited} entities (e.g.\ various concrete objects) when
the relevant participant remains constant across the continuum.  For
example, with our pouring process ($e$), the agent (Jack) and the goal
(bucket A) remain constant across the particular events which make up
the continuum.  In contrast, the patient (varying quantities of water)
does not remain constant; for this reason, a substance ($x$) must be
used with the patient role in order to satisfy the above principle.
As a corollary, note that if a particular quantity of water (e.g.\
$x_1$) is supplied for the patient role instead, process predication
becomes impossible, which forces the predication to be over an event
(e.g.\ $e_1$).  While space precludes further discussion here, this
brief sketch should indicate how the conventional wisdom concerning
{\em set terminal points\/} can be realized along sortal lines.

To conclude this section, I shall briefly compare the present account
to two other related ones.  First, in its use of abstract entities
(the continua) whose elements (or {\em realizations\/}) vary in
amount, the present account is reminiscent of
Hinrichs~\shortcite{Hinrichs85}, where Carlsonian kinds are employed.
In contrast to the present approach though, Hinrichs makes essentially
no use of the sortal distinctions he proposes in the eventuality
domain; moreover, unlike Krifka and Jackendoff, he does not propose a
uniform treatment of adjectival and adverbial measure
phrases.\footnote{Yet another alternative would be to follow Hinrichs
in his use of kinds in the nominal domain only, while abandonding his
quantificational analysis of {\em for}-adverbials in favor of an
analysis more in the spirit of Krifka's approach.  To successfully
develop such an approach, however, requires several tricky issues
concerning the scope of indefinites to be resolved.  These issues are
currently under investigation (in cooperation with Alessandro
Zucchi).} Second, the present account is also very much in the spirit
of Verkuyl~\shortcite{Verkuyl93}, especially in its attention to the
problems posed by non-individuating accomplishment expressions.  As
the next section is intended to show, however, its sortal basis
appears to make it better suited for computational purposes.

\section{A Calculus of Eventualities}

To illustrate how the present approach to aspectual composition
maintains the conciseness advantage of eventuality-based knowledge
representations cited previously, let us now consider two examples in
some detail.

First, suppose our knowledge base contains the first representation of
the event $e_1$ given near the end of the preceding section, repeated
below:

{\singlespace
{\small
\[
\begin{array}{c}
\left[
\begin{array}{l}
{\sf index}: \; e_1 \\
{\sf sort}: \; {\sf event} \\
{\sf composed\mbox{-}of}: \;
  \left[
  \begin{array}{l}
  {\sf index}: \; e \\
  {\sf sort}: \; {\sf process} \\
  {\sf pred}: \; {\sf pour} \\
  {\sf agent}: \; {\sf jack} \\
  {\sf patient}: \;
    \left[
    \begin{array}{l}
    {\sf index}: \; x \\
    {\sf sort}: \; {\sf substance} \\
    {\sf pred}: \; {\sf water} \\
    \end{array}
    \right] \\
  {\sf goal}: \;
    \left[
    \begin{array}{l}
    {\sf index}: \; x_a \\
    {\sf sort}: \; {\sf object} \\
    {\sf pred}: \; {\sf bucket} \\
    {\sf name}: \; {\sf A} \\
    \end{array}
    \right] \\
  \end{array}
  \right] \\
{\sf duration}:
  \left[
  \begin{array}{l}
  {\sf number}: 30 \\
  {\sf unit}: \; {\sf seconds} \\
  \end{array}
  \right] \\
\end{array}
\right] \\
\mbox{} \\
\mbox{\em Jack poured water into bucket A for thirty seconds} \\
\end{array}
\]
}}

\noindent From this representation of $e_1$, we should be able to derive the
existence of an event $e_2$ in which Jack pours water into bucket A
for twenty-five seconds, as well as an event $e_3$ in which he does so
for twenty seconds, and so forth.  This can be achieved using the
following rule:

{\singlespace
{\small
\[
\begin{array}{c}
\left[
\begin{array}{l}
{\sf index}: \; E_1 \\
{\sf sort}: \; {\sf event} \\
{\sf composed\mbox{-}of}: \; E \\
{\sf duration}:
  \left[
  \begin{array}{l}
  {\sf number}: N_1 \\
  {\sf unit}: \; U \\
  \end{array}
  \right] \\
\end{array}
\right] \\
\mbox{} \\
\mbox{\em $E$ for $N_1$ $U$s} \\
\end{array}
\; \; \Longrightarrow \; \;
\begin{array}{c}
\left[
\begin{array}{l}
{\sf index}: \; E_2 \\
{\sf sort}: \; {\sf event} \\
{\sf composed\mbox{-}of}: \; E \\
{\sf duration}:
  \left[
  \begin{array}{l}
  {\sf number}: N_2 \\
  {\sf unit}: \; U \\
  \end{array}
  \right] \\
\end{array}
\right] \\
\mbox{} \\
\mbox{\em $E$ for $N_2$ $U$s} \\
\end{array}
\]
\[
{\rm where} \; N_2 \; \leq \; N_1
\]
}}

This rule reflects a pair of assumptions regarding processes (i.e.,
process continua) such as $e$.  First, it is assumed that the
particular events making up the continuum $e$ are closed under
the subpart relation; thus, if the event $e_1$ composed-of $e$
has a subevent $e_2$, then $e_2$ must also be composed-of $e$.
Second, if $e_1$ has duration $N_1$ $U$s, then for all non-negative
numbers $N_2$ less than $N_1$, $e_1$ is assumed to have a subevent $e_2$
of duration $N_2$ $U$s (clearly a simplifying assumption!).  Taken
together, these two assumptions yield the above rule as a theorem.

Second, let us now suppose that we have a method for calculating the
rate of transfer for the pouring process $e$ above, and that this rate
multiplied by thirty seconds turns out to be five gallons.  Using
this information, we should then be able to derive the second
representation of the event $e_1$ given near the end of the preceding
section (repeated below) from the first one.

{\singlespace
{\small
\[
\begin{array}{c}
\left[
\begin{array}{l}
{\sf index}: \; e_1 \\
{\sf sort}: \; {\sf event} \\
{\sf pred}: \; {\sf pour} \\
{\sf agent}: \; {\sf jack} \\
{\sf patient}: \;
  \left[
  \begin{array}{l}
  {\sf index}: \; x_1 \\
  {\sf sort}: \; {\sf object} \\
  {\sf composed\mbox{-}of}: \;
    \left[
    \begin{array}{l}
    {\sf index}: \; x \\
    {\sf sort}: \; {\sf substance} \\
    {\sf pred}: \; {\sf water} \\
    \end{array}
    \right] \\
  {\sf quantity}: \;
    \left[
    \begin{array}{l}
    {\sf number}: \; 5 \\
    {\sf unit}: \; {\sf gallons} \\
    \end{array}
    \right] \\
  \end{array}
  \right] \\
{\sf goal}: \;
  \left[
  \begin{array}{l}
  {\sf index}: \; x_a \\
  {\sf sort}: \; {\sf object} \\
  {\sf pred}: \; {\sf bucket} \\
  {\sf name}: \; {\sf A} \\
  \end{array}
  \right] \\
{\sf duration}:
  \left[
  \begin{array}{l}
  {\sf number}: 30 \\
  {\sf unit}: \; {\sf seconds} \\
  \end{array}
  \right] \\
\end{array}
\right] \\
\mbox{} \\
\mbox{\em Jack poured five gallons of water into bucket A
in thirty seconds} \\
\end{array}
\]
}}

\noindent This can be achieved using the following rule:

{\singlespace
{\footnotesize
\[
\begin{array}{c}
\left[
\begin{array}{l}
{\sf index}: \; E_1 \\
{\sf sort}: \; {\sf event} \\
{\sf composed\mbox{-}of}: \;
  \left[
  \begin{array}{l}
  {\sf index}: \; E \\
  {\sf sort}: \; {\sf process} \\
  {\sf pred}: \; {\sf pour} \\
  {\sf agent}: \; A \\
  {\sf patient}: X \\
  {\sf goal}: \; G \\
  \end{array}
  \right] \\
{\sf duration}:
  \left[
  \begin{array}{l}
  {\sf number}: N_1 \\
  {\sf unit}: \; U_1 \\
  \end{array}
  \right] \\
\end{array}
\right] \\
\mbox{} \\
\mbox{\em $A$ pours $X$ into $G$ for $N_1$ $U_1$s} \\
\end{array}
\; \; \Longrightarrow \; \;
\begin{array}{c}
\left[
\begin{array}{l}
{\sf index}: \; E_1 \\
{\sf sort}: \; {\sf event} \\
{\sf pred}: \; {\sf pour} \\
{\sf agent}: \; A \\
{\sf patient}: \;
  \left[
  \begin{array}{l}
  {\sf index}: \; X_1 \\
  {\sf sort}: \; {\sf object} \\
  {\sf composed\mbox{-}of}: \; X \\
  {\sf quantity}: \;
    \left[
    \begin{array}{l}
    {\sf number}: \; N_2 \\
    {\sf unit}: \; U_2 \\
    \end{array}
    \right] \\
  \end{array}
  \right] \\
{\sf goal}: \; G \\
{\sf duration}:
  \left[
  \begin{array}{l}
  {\sf number}: N_1 \\
  {\sf unit}: \; U_1 \\
  \end{array}
  \right] \\
\end{array}
\right] \\
\mbox{} \\
\mbox{\em $A$ pours $N_2$ $U_2$s of $X$ into $G$ in $N_1$ $U_1$s} \\
\end{array}
\]
\[
{\rm where} \; N_2 \; = \; {\sf rate}(E, U_2, U_1) \; \times \; N_1
\]
}}

\noindent Since this rule may look overly complicated at first glance,
let us pause to consider why it makes sense.  Because the process $E$
of which the event $E_1$ is composed is understood to be a continuum
of pouring events, all with agent $A$ and goal $G$, we can derive that
$E_1$ in particular is a pouring event with agent $A$ and goal $G$.
As for the patient of $E_1$, we may note that since the patient of $E$
is understood to the continuum $X$ of quantities poured, one of these
must be the patient of $E_1$; supposing that its index is $X_1$, we
can then derive that $X_1$ is both the patient of $E_1$ and composed
of $X$.  (The calculation of $X_1$'s amount, $N_2$ $U_2$s, is not of
particular concern here, and thus is assumed to be straightforward.)

In the actual implementation, this rule has been generalized to handle
other verbs which form incremental thematic relations, such as {\em
dribble, drip, leak, ooze, seep, siphon\/} and so on.  Several
additional cases have also been implemented, including the
progressive, {\em at}-adverbials and the aspectual verbs {\em start,
stop\/} and {\em finish}.  These rules are described in detail in
\cite{White94Thesis}.

\section{Space and Time Linked}

The present account extends naturally to cover motion verbs, which are
assumed to form incremental thematic relations between paths and
eventualities (rather than between material entities and
eventualities).  For the most part, all that is required is to
introduce the appropriate counterparts of the substances and processes
into the domain of directed spatial entities.

Following Krifka, I will assume that events of directed motion have
spatial traces as well as temporal ones, i.e., that each directed
motion event has a unique (delimited) path and time interval
associated with it.  With directed motion processes, however, this
cannot be the case, given the present conception of predication over
continua: while a continuum of motion events may have the same agent
and the same manner of motion, the path traversed does not remain
constant.  Consequently, if a directed motion process is to have a
unique path associated with it, the path must likewise be a continuum,
i.e.\ what I shall call a {\bf non-delimited path}.

Once the notion of incremental thematic relation has been extended to
the case of directed motion eventualities in this way, it remains only
to examine the sortal restrictions which make sense for various path
predicates.  For example, with {\em to}-phrases (e.g.\ {\em to the
bridge}), which specify endpoints, we may naturally assume that their
translations are only well-sorted with delimited paths, since
endpoints do not remain constant across a continuum of such paths.  As
a result, expressions such as {\em Jack run to the bridge\/} will give
rise to predicates restricted to events, in contrast to {\em Jack
run\/} (which can apply to processes as well).  This explains why {\em
* Jack ran to the bridge for thirty seconds\/} is not well-formed, as
illustrated below (recall that the composed-of relation serves to map
processes to events):

{\singlespace
{\small
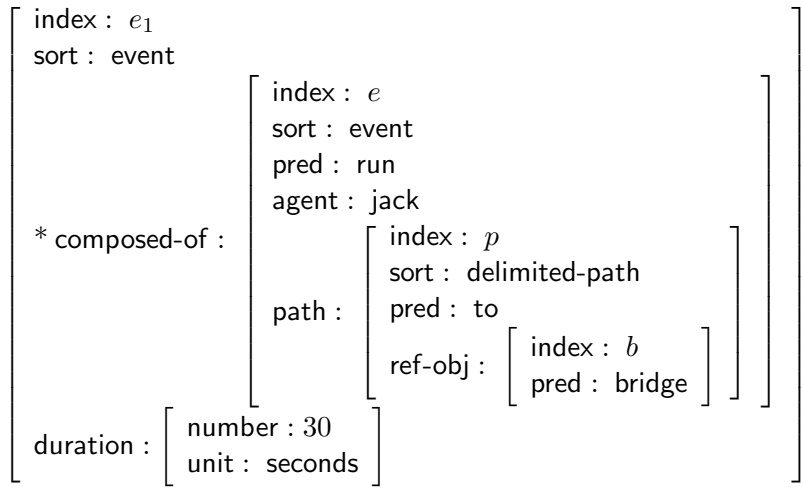
\[
\begin{array}{c}
\left[
\begin{array}{l}
{\sf index}: \; e_1 \\
{\sf sort}: \; {\sf event} \\
\mbox{*} \: {\sf composed\mbox{-}of}: \;
  \left[
  \begin{array}{l}
  {\sf index}: \; e \\
  {\sf sort}: \; {\sf event} \\
  {\sf pred}: \; {\sf run} \\
  {\sf agent}: \; {\sf jack} \\
  {\sf path}: \;
    \left[
    \begin{array}{l}
    {\sf index}: \; p \\
    {\sf sort}: \; {\sf delimited\mbox{-}path} \\
    {\sf pred}: \; {\sf to} \\
    {\sf ref\mbox{-}obj}: \;
      \left[
      \begin{array}{l}
      {\sf index}: \; b \\
      {\sf pred}: \; {\sf bridge} \\
      \end{array}
      \right]
    \end{array}
    \right] \\
  \end{array}
  \right] \\
{\sf duration}:
  \left[
  \begin{array}{l}
  {\sf number}: 30 \\
  {\sf unit}: \; {\sf seconds} \\
  \end{array}
  \right] \\
\end{array}
\right] \\
\mbox{} \\
\mbox{\em * Jack ran to the bridge for thirty seconds} \\
\end{array}
\]
}}

Unlike {\em to}-phrases, {\em towards}-phrases do make sense for
non-delimited paths, since these specify direction rather than
endpoints (and direction can remains constant across a continuum).
As such, {\em Jack ran towards the bridge for thirty seconds\/} receives
the following well-formed translation:

{\singlespace
{\small
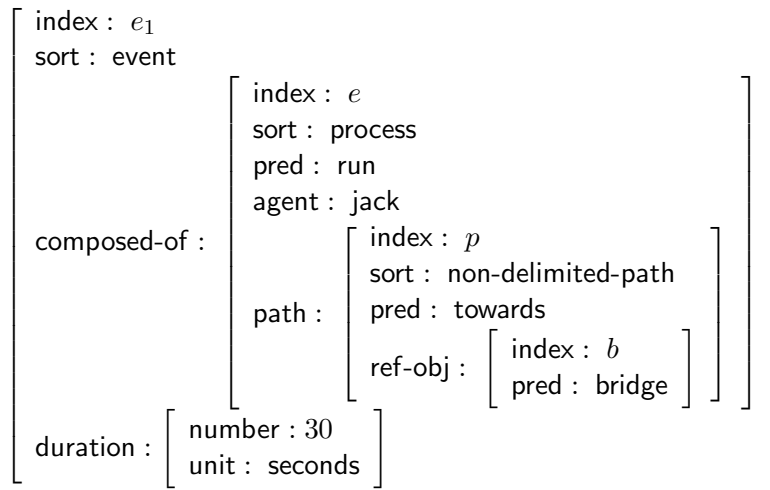
\[
\begin{array}{c}
\left[
\begin{array}{l}
{\sf index}: \; e_1 \\
{\sf sort}: \; {\sf event} \\
{\sf composed\mbox{-}of}: \;
  \left[
  \begin{array}{l}
  {\sf index}: \; e \\
  {\sf sort}: \; {\sf process} \\
  {\sf pred}: \; {\sf run} \\
  {\sf agent}: \; {\sf jack} \\
  {\sf path}: \;
    \left[
    \begin{array}{l}
    {\sf index}: \; p \\
    {\sf sort}: \; {\sf non\mbox{-}delimited\mbox{-}path} \\
    {\sf pred}: \; {\sf towards} \\
    {\sf ref\mbox{-}obj}: \;
      \left[
      \begin{array}{l}
      {\sf index}: \; b \\
      {\sf pred}: \; {\sf bridge} \\
      \end{array}
      \right]
    \end{array}
    \right] \\
  \end{array}
  \right] \\
{\sf duration}:
  \left[
  \begin{array}{l}
  {\sf number}: 30 \\
  {\sf unit}: \; {\sf seconds} \\
  \end{array}
  \right] \\
\end{array}
\right] \\
\mbox{} \\
\mbox{\em Jack ran towards the bridge for thirty seconds} \\
\end{array}
\]
}}

\noindent
As an aside, it is worth observing that non-delimited paths need not
be {\em unbounded}.  This is especially important with predicates such
as {\sf towards}, since the reference object here serves to impose an
upper limit on how far the continuum can extend.  For example,
consider the process $e$ above of Jack running towards the bridge
(assumed to be of more or less constant speed and direction).
Although the continuum $e$ may contain events larger than $e_1$, it
cannot contain any events larger than the event in which Jack reaches
the bridge, as the path of any such event would no longer satisfy the
predicated yielded by {\em towards the bridge}.  Because the present
notion of delimitedness is independent of boundedness (in the
mathematical sense), the presence of upper bounds in cases such as
this one is entirely unproblematic.

Another interesting case is that of {\em along}.  In general, distance
cannot be predicated of a non-delimited path, as distance varies
according to the endpoints.  This is not the case, however, with
{\bf proximal distance}, i.e.\ the distance between the path and
the reference object, which can remain constant across a path
continuum.  This explains why sentences like {\em Jack ran along
the river, two hundred yards from the shore, for thirty seconds}
should be well-formed.

Finally we turn to distance phrases.  As mentioned above, distance
cannot be sensibly predicated of non-delimited paths, so I will assume
that distance predication is restricted to delimited paths.
Consequently, bare distance phrases will behave just like {\em
to}-PPs, which explains both why {\em Jack ran two miles to the
bridge\/} is fine and why {\em * Jack ran two miles for ten minutes\/}
is out.  Now, what about distance phrases headed by {\em for}?
Remarkably, these adverbials have been almost completely ignored in
the literature.  As with their temporal counterparts, I will assume
that distance {\em for}-adverbials form measure phrases, i.e.\
serve to introduce the composed-of mapping.  By making this natural
assumption, we may then explain the curious fact that {\em *~Jack ran to the
bridge for two miles\/} is horrible, in sharp contrast to both {\em Jack
ran two miles to the bridge\/} and {\em Jack ran along the river
for two miles.}  This is illustrated below:

{\singlespace
{\small
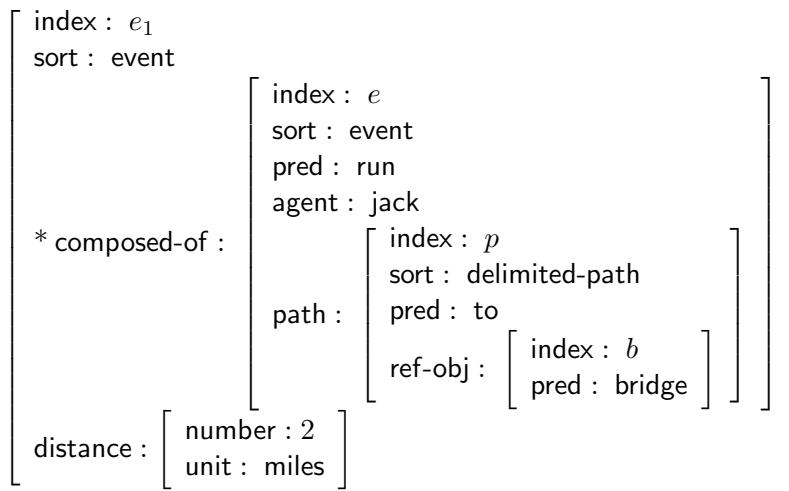
\[
\begin{array}{c}
\left[
\begin{array}{l}
{\sf index}: \; e_1 \\
{\sf sort}: \; {\sf event} \\
\mbox{*} \: {\sf composed\mbox{-}of}: \;
  \left[
  \begin{array}{l}
  {\sf index}: \; e \\
  {\sf sort}: \; {\sf event} \\
  {\sf pred}: \; {\sf run} \\
  {\sf agent}: \; {\sf jack} \\
  {\sf path}: \;
    \left[
    \begin{array}{l}
    {\sf index}: \; p \\
    {\sf sort}: \; {\sf delimited\mbox{-}path} \\
    {\sf pred}: \; {\sf to} \\
    {\sf ref\mbox{-}obj}: \;
      \left[
      \begin{array}{l}
      {\sf index}: \; b \\
      {\sf pred}: \; {\sf bridge} \\
      \end{array}
      \right]
    \end{array}
    \right] \\
  \end{array}
  \right] \\
{\sf distance}:
  \left[
  \begin{array}{l}
  {\sf number}: 2 \\
  {\sf unit}: \; {\sf miles} \\
  \end{array}
  \right] \\
\end{array}
\right] \\
\mbox{} \\
\mbox{\em * Jack ran to the bridge for two miles} \\
\end{array}
\]
}}

{\singlespace
{\small
\[
\begin{array}{c}
\left[
\begin{array}{l}
{\sf index}: \; e_1 \\
{\sf sort}: \; {\sf event} \\

  {\sf pred}: \; {\sf run} \\
  {\sf agent}: \; {\sf jack} \\
  {\sf path}: \;
    \left[
    \begin{array}{l}
    {\sf index}: \; p_1 \\
    {\sf sort}: \; {\sf delimited\mbox{-}path} \\
    {\sf pred}: \; {\sf to} \\
    {\sf ref\mbox{-}obj}: \;
      \left[
      \begin{array}{l}
      {\sf index}: \; b \\
      {\sf pred}: \; {\sf bridge} \\
      \end{array}
      \right]
    \end{array}
    \right] \\

{\sf distance}:
  \left[
  \begin{array}{l}
  {\sf number}: 2 \\
  {\sf unit}: \; {\sf miles} \\
  \end{array}
  \right] \\
\end{array}
\right] \\
\mbox{} \\
\mbox{\em Jack ran two miles to the bridge} \\
\end{array}
\]
}}

{\singlespace
{\small
\[
\begin{array}{c}
\left[
\begin{array}{l}
{\sf index}: \; e_1 \\
{\sf sort}: \; {\sf event} \\
{\sf composed\mbox{-}of}: \;
  \left[
  \begin{array}{l}
  {\sf index}: \; e \\
  {\sf sort}: \; {\sf process} \\
  {\sf pred}: \; {\sf run} \\
  {\sf agent}: \; {\sf jack} \\
  {\sf path}: \;
    \left[
    \begin{array}{l}
    {\sf index}: \; p \\
    {\sf sort}: \; {\sf non\mbox{-}delimited\mbox{-}path} \\
    {\sf pred}: \; {\sf along} \\
    {\sf ref\mbox{-}obj}: \;
      \left[
      \begin{array}{l}
      {\sf index}: \; r \\
      {\sf pred}: \; {\sf river} \\
      \end{array}
      \right]
    \end{array}
    \right] \\
  \end{array}
  \right] \\
{\sf distance}:
  \left[
  \begin{array}{l}
  {\sf number}: 2 \\
  {\sf unit}: \; {\sf miles} \\
  \end{array}
  \right] \\
\end{array}
\right] \\
\mbox{} \\
\mbox{\em Jack ran along the river for two miles} \\
\end{array}
\]
}}

\noindent
One consequence of the present approach is that {\em Jack run along
the river for two miles\/} and {\em Jack run two miles along the
river\/} are assigned rather different representations.  Although
space precludes further discussion here, this should not be considered
particularly troublesome since these sentences can easily be made to
be mutually entailing.

\section{Conclusion}

In this paper, I have argued that a sortal approach to aspectual
composition, developed in the spirit of the eventuality-based work on
episodic representation, can indeed be explanatory.  In so doing, I
have argued against the prevailing opinion in the linguistics and
philosophy literature that such approaches inherently represent a dead
end.  In support of this view, I have developed a synthesis of
competing proposals by Hinrichs~\shortcite{Hinrichs85},
Krifka~\shortcite{Krifka92} and Jackendoff~\shortcite{Jackendoff91}
which takes Jackendoff's cross-cutting sortal distinctions as its
point of departure (cf. \cite{White94Thesis} for further details).
Nevertheless, many other possible ways of implementing a sortal
approach remain to be explored, as do many empirical issues, and thus
the question of whether a sortal approach to aspectual composition
is to be preferred is likely to remain open for some time.

\nocite{Carlson77Thesis}
\nocite{Dowty79}
\nocite{Kent93}
\nocite{Parsons90}
\nocite{Pustejovsky91Cognition}
\nocite{SchubertPelletier87}
\nocite{Vendler67}
\nocite{WhiteEACL93}
\nocite{WhiteACLStudent93}
\nocite{WhiteAAAI94}



\end{document}